\documentclass[10pt,twocolumn,letterpaper]{article}

\usepackage{cvpr}
\usepackage{graphicx}
\usepackage{times}
\usepackage{amsmath}
\usepackage{amssymb}

\usepackage[utf8]{inputenc} 
\usepackage[T1]{fontenc}    
\usepackage{url}            
\usepackage{booktabs}       
\usepackage{amsfonts}       
\usepackage{nicefrac}       
\usepackage{microtype}      
\usepackage{appendix}
\usepackage{amsthm}
\usepackage{subcaption}
\usepackage{bm}
\usepackage{mathtools}

\usepackage{tikz}
\usepackage{mathdots}
\usepackage{yhmath}
\usepackage{cancel}
\usepackage{color}
\usepackage{siunitx}
\usepackage{array}
\usepackage{multirow}
\usepackage{tabularx}
\usepackage{booktabs}
\usetikzlibrary{fadings}
\usepackage{gensymb}

\usepackage{algorithm}
\usepackage{algorithmic}

\graphicspath{{fig/}}

\newcommand{\trainset}[1]{\mathcal{S}^{#1}}
\newcommand{\visual}[1]{\mathcal{V}^{#1}}
\newcommand{\textual}[1]{\mathcal{T}^{#1}}
\newcommand{\Z}[1]{\mathcal{Z}^{#1}}
\newcommand{\Y}[2]{\mathcal{Y}^{#1}_{#2}}
\renewcommand{\vec}[1]{\mathbf{#1}}
\newcommand{\metric}{d}
\newcommand{\Proba}{\mathbb{P}}

\renewcommand{\vec}[1]{\mathbf{#1}}

\DeclareMathOperator{\softmax}{softmax}
\DeclareMathOperator{\crossentropy}{crossentropy}

\usepackage[pagebackref=true,breaklinks=true,letterpaper=true,colorlinks,bookmarks=false]{hyperref}

\cvprfinalcopy 

\def\cvprPaperID{16} 

\ifcvprfinal\fi
\begin{document}

\title{CLAREL: Classification via retrieval loss for zero-shot learning}


\author{%
    Boris N. Oreshkin \\
    Element AI \\
    \texttt{boris.oreshkin@gmail.com} \\
    \and
    Negar Rostamzadeh \\
    Element AI \\
    \texttt{negar@elementai.com}
    \and
    Pedro O. Pinheiro \\
    Element AI \\
    \texttt{pedro@elementai.com} \\
    \and
    Christopher Pal \\
    Element AI \\
    \texttt{christopher.pal@elementai.com} \\
}

\maketitle

\begin{abstract}
   We address the problem of learning cross-modal representations. We propose an instance-based deep metric learning approach in joint visual and textual space. The key novelty of this paper is that it shows that using per-image semantic supervision leads to substantial improvement in zero-shot performance over using class-only supervision. We also provide a probabilistic justification and empirical validation for a metric rescaling approach to balance the seen/unseen accuracy in the GZSL task. We evaluate our approach on two fine-grained zero-shot datasets: \textsc{cub} and \textsc{flowers}.
\end{abstract}

\section{Introduction}

Deep learning-based approaches have demonstrated superior flexibility and generalization capabilities in information processing on a wide variety of tasks, such as vision, speech and language~\cite{lecun2015nature}. However, it has been widely realized that the transfer of deep representations to real-world applications is challenging due to the typical reliance on massive hand-labeled datasets. Learning in the low-labeled data regime, especially in the zero-shot~\cite{wang2019asurvey} and the few-shot~\cite{wang2019fewshot} setups, have recently received significant attention in the literature. In the problem of zero-shot learning (ZSL), the objective is to recognize categories that have not been seen during the training~\cite{larochelle2008zsl} via modality alignment. This is an especially relevant problem as machine learning is challenged with the long tail of classes, and the idea of learning from pairs of images and sentences, abundant on the web, looks like a natural solution. Therefore, in this paper we specifically target the fine-grained scenario of paired images and their respective text descriptions. The uniqueness of this scenario is in the fact that the co-occurance of image and text provides a rich source of information. The ways of leveraging this source have not been sufficiently explored in the context of ZSL. 

In this paper, we specifically target the fine-grained visual description scenario, as defined by Reed et al.~\cite{reed16learning}. Concretely, given a training set $\trainset{} = \{ (v_n, t_n, y_n)\;|\; v_n \in \visual{},\ t_n \in \textual{}, y_n \in \Y{}{}, n = 1\ldots N \}$ of image, text and label tuples, we are interested in finding representations $f_{\phi} : \visual{} \rightarrow \Z{}$ of image, parameterized by $\phi$, and $f_{\theta} : \textual{} \rightarrow \Z{}$ of text, parameterized by $\theta$, in a common embedding space $\Z{}$. Furthermore, generalized ZSL (GZSL) problem is defined using the sets of seen $\Y{tr}{}$ and unseen $\Y{ts}{}$ classes, such that $\Y{}{} = \Y{tr}{} \cup \Y{ts}{}$ and $\Y{tr}{} \cap \Y{ts}{} = \emptyset$. The training set only contains the seen classes, \emph{i.e.} $\trainset{tr} = \{ (v_n, t_n, y_n)\;|\;v_n \in \visual{},\ t_n \in \textual{}, y_n \in \Y{tr}{}\}$ and the task is to build a classifier function $g : \Z{} \times \Z{} \rightarrow \Y{}{}$.
This is different from the ZSL scenario focusing on $g : \Z{} \times \Z{} \rightarrow \Y{ts}{}$. The most acute problem in GZSL setup is the accuracy imbalance between seen and unseen classes. To measure and control the imbalance, three metrics are commonly used to assess the classification performance in the GZSL scenario: the Top-1 accuracy on the seen categories (\textbf{s}), the Top-1 accuracy on the unseen categories (\textbf{u}) and their harmonic mean, $\textbf{H} = \textbf{u} \cdot \textbf{s} / (\textbf{u} + \textbf{s})$. The contributions of this work can be characterized under the following two themes.

\textbf{Instance-based training loss.} Zero-shot learning approaches rely heavily on class-level modality alignment~\cite{xian2018zero}. We propose a new composite loss function that balances instance-based pairwise image/text retrieval loss and the usual classifier loss. The retrieval loss term does not use class labels. We show that most of the GZSL accuracy can be extracted from the instance-based retrieval loss.

\textbf{Metric rescaling.} GZSL approaches suffer from imbalanced performance on seen and unseen classes~\cite{shichen2018generalized}. Previous work proposed to use a heuristic trick, calibrated stacking~\cite{chao2016anempirical} or calibration~\cite{das2019zero}, to solve the problem. We provide a sound probabilistic justification for it. 

\begin{figure*}[t]
    \centering
    \resizebox {.8\textwidth} {!} {
        \input{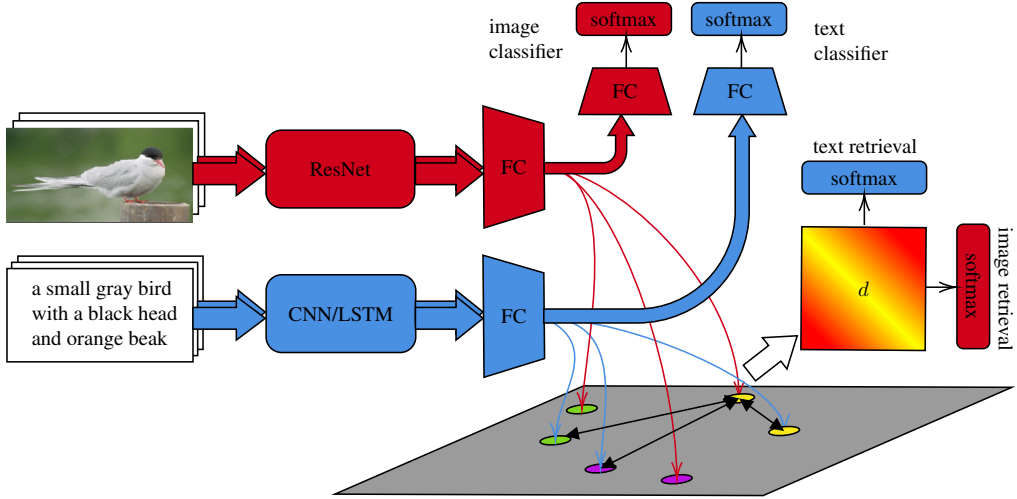}
    }
    \caption{Proposed method. Each batch consists of randomly sampled instances, \emph{i.e.} pairs of images and their corresponding texts. Images are embedded via ResNet and texts are embedded via a CNN/LSTM stack. Image and text features are projected via a fully connected layer into the same dimensional space. The negative distances between text and image features are fed into softmax to train on the image and the text retrieval tasks. In addition, image and text embeddings are trained on auxiliary image and text classification tasks on the class labels corresponding to instances.}
    \label{fig:architecture}
\end{figure*}

\section{Proposed Method}

To build $g$, most approaches to joint representation learning rely on class labeling to train a representation. For example, all the methods reviewed by Xian et al.~\cite{xian2018zero} require the access to class labels at train time. We hypothesise that in the fine-grained learning scenario, such as the one described by Reed et al.~\cite{reed16learning}, a lot of information can be extracted simply from pairwise image/text co-occurrences. The class labels really only become necessary when we define class prototypes, \emph{i.e.} at zero-shot test time. Following this intuition, we define a framework based on projecting texts and images into a common space and then learning a representation based on a mixture of four loss functions: a pairwise text retrieval loss, a pairwise image retrieval loss, a text classifier loss and an image classifier loss (see Fig.~\ref{fig:architecture} and Algorithm~\ref{alg:one_iteration} in Appendix~\ref{sec:loss_algorithm}). The framework enables us, among other things, to experiment with the effects of train-time availability of class labels on the quality of zero-shot representations.

\textbf{Pairwise cross-modal loss function} is based solely on the pairwise relationships between texts and images. Suppose $d$ is a metric $\metric : \Z{} \times \Z{} \rightarrow \mathbb{R}^+$, $v_i$ is an image and $\tau = \{ t_{j^\prime} \}$ is a collection of arbitrary texts sampled uniformly at random, of which text $t_j$ belongs to $v_i$. We propose the following model for the probability of image $v_i$ and text $t_j$ to belong to the same object instance: 
\begin{align*}
p_{\phi,\theta}(i=j|v_i, t_j, \tau) = \frac{\exp(-\metric(f_{\phi}(v_i), f_{\theta}(t_j)))}{\sum_{t_{j^\prime} \in \tau} \exp(-\metric(f_{\phi}(v_i), f_{\theta}(t_{j^\prime})))}\;.
\end{align*}
The learning is then based on the cross-entropy log-loss defined on the batch of size $B$:
\begin{align*} 
    J_{TR}(\phi,\theta) &= - \frac{1}{B} \sum_{i,j = 1}^{B} \ell_{i,j} \log p_{\phi,\theta}(i=j|v_i, t_j, \{t_{j^\prime}\}_{j^\prime=1}^B),
\end{align*}
where $\ell_{i,j}$ is a binary indicator of the true match ($\ell_{i,j}=1 \textrm{, if } i=j \textrm{ and } 0 \textrm{ otherwise}$). Note that the expression above has the interpretation of the text retrieval loss. Exchanging the order of image and text in the probability model $J_{TR}(\phi,\theta)$ leads to the image retrieval loss, $J_{IR}(\phi,\theta)$. The two losses are mixed using parameter $\lambda \in [0, 1]$ as shown in Algorithm~\ref{alg:one_iteration} in Appendix~\ref{sec:loss_algorithm}. The pairwise retrieval loss functions are responsible for the modality alignment. In addition to those, we propose to include the usual image and text classifier losses responsible for reducing the intraclass variability of representations. The classifier losses are added to the retrieval losses using a mixing parameter $\kappa \in [0,1]$ as shown in Algorithm~\ref{alg:one_iteration} in Appendix~\ref{sec:loss_algorithm}.

\subsection{Balancing Accuracy for Seen and Unseen} \label{ssec:seen_unseen_balancing_theory}

Let us define class prototype $\vec{p}(y)$ based on the set of texts $\textual{}_y$ belonging to class $y$, $\vec{p}(y) = \frac{1}{|\textual{}_y|} \sum_{t_i \in \textual{}_y} f_{\theta}(t_i)$. In GZSL, the nearest neighbor decision rule for a given image $v$ and its features $\vec{z}_v = f_{\phi}(v)$ has the following form:
\begin{align} \label{eqn:nn_classifier}
\widehat y = \arg\min_{y \in \Y{}{}} d(\vec{z}_v, \vec{p}(y))\;.
\end{align}

To formalize the problem, we first introduce $y_v$, the true class label of image $v$. Mathematically, the main GZSL pain point is that $\Proba\{ \widehat y \in \Y{tr}{} | y_v \in \Y{ts}{} \}$ is significantly greater than $\Proba\{ \widehat y \in \Y{ts}{} | y_v \in \Y{tr}{} \}$. In other words, the problem is that a given image is more likely to be confused with one of the seen classes if it belongs to an unseen class than vice versa. We propose the following probabilistic representation of the event space for the decision rule in Equation~\eqref{eqn:nn_classifier}:
\begin{align} \label{eqn:seen_unseen_probability_representation}
\Proba\{ \widehat y \in \Y{tr}{} | y_v \in &\Y{ts}{} \} = \Proba\left\{\min_{y \in \Y{tr}{}} d(\vec{z}_v, \vec{p}(y)) \right. \nonumber \\
&< \left. \min_{y \in \Y{ts}{}} d(\vec{z}_v, \vec{p}(y)) \;|\; y_v \in \Y{ts}{} \right\}\;.
\end{align}
To balance $\Proba\{ \widehat y \in \Y{tr}{} | y_v \in \Y{ts}{} \}$ and $\Proba\{ \widehat y \in \Y{ts}{} | y_v \in \Y{tr}{} \}$, we introduce a positive scalar $\alpha \in \mathbb{R}^+$ and scale all the distances corresponding to the seen prototypes by $1+\alpha$, giving rise to the scaled distance $d_{\alpha}$:
\begin{align*} 
    d_{\alpha}(\vec{z}_v, \vec{p}(y))= 
    \begin{cases}
        (1+\alpha)d(\vec{z}_v, \vec{p}(y)), & \text{if } y \in \Y{tr}{} \\
        d(\vec{z}_v, \vec{p}(y)), & \text{otherwise}
    \end{cases}\;
\end{align*}
The error probability classifying unseen classes as seen ones for the classifier based on $d_{\alpha}(\vec{z}_v, \vec{p}(y))$, $\Proba\{ \widehat y_{\alpha} \in \Y{tr}{} | y_v \in \Y{ts}{} \}$, is then a monotone non-increasing function of $\alpha$ and we can reduce it by increasing $\alpha$ (please refer to Appendix~\ref{sec:proof} for a proof). Consider now $\Proba\{ \widehat y_{\alpha} \in \Y{tr}{} | y_v \in \Y{tr}{} \}$, which is a probability that we classify an image $v$ from one of the seen classes as still one of the seen classes. Using exactly the same chain of arguments as in Appendix~\ref{sec:proof} we can show that the probability is a non-increasing function of $\alpha$. Hence the probability $\Proba\{ \widehat y_{\alpha} \in \Y{ts}{} | y_v \in \Y{tr}{} \} = 1 - \Proba\{ \widehat y_{\alpha} \in \Y{tr}{} | y_v \in \Y{tr}{} \}$ is a non-decreasing function of $\alpha$. Therefore, we expect that by varying $\alpha > 0$ we can balance the error rates $\Proba\{\widehat y_{\alpha} \in \Y{tr}{} | y_v \in \Y{ts}{} \}$ and $\Proba\{\widehat y_{\alpha} \in \Y{ts}{} | y_v \in \Y{tr}{} \}$. 

\begin{table}[t]
    \centering
    \caption{Generalized zero-shot Top-1 classification accuracy.}
    \label{table:key_empirical_result_gzsl}
    \begin{tabular}{lcccccc} 
        \toprule
         & \multicolumn{3}{c}{\textsc{cub}}  & \multicolumn{3}{c}{\textsc{flowers}} \\ 
         & \textbf{u} &  \textbf{s} &  \textbf{H}  & \textbf{u} &  \textbf{s} &  \textbf{H} \\ 
        \hline
        \textsc{cada-vae}~\cite{schonfeld2018generalized}  &  n/a & n/a & 53.4 & n/a & n/a & n/a  \\
        Xian et al.~\cite{xian2018feature} & 50.3 & 58.3 & 54.0  & 59.0 & 73.8 & 65.6 \\
        Xian et al.~\cite{xian2019fVAEGAND2} & 48.4 & 60.1 & 53.6 & 56.8 & 74.9 & 64.6 \\
        Felix et al.~\cite{rafael2018multimodal} & 47.9 & 59.3 & 53.0 & 61.6 & 69.2 & 65.2 \\
        Atzmon et al.\cite{atzmon2019adaptive} & n/a & n/a & n/a & 59.6 & 81.4  & 68.8\\
        \hline
        Ours & 59.3 & 52.6 & \textbf{55.8} & 73.0 & 73.6 & \textbf{73.3}  \\
        \bottomrule
    \end{tabular}
\end{table}

\begin{table}[t]
    \centering
    \caption{Zero-shot Top-1 classification accuracy.}
    \label{table:key_empirical_result_zsl}
    \begin{tabular}{lccr} 
        \toprule
         & \multicolumn{1}{c}{\textsc{cub}}  & \multicolumn{1}{c}{\textsc{flowers}} &  \\ 
        \hline
        f-CLSWGAN~\cite{xian2018feature} & 57.3 & 67.2  \\
        f-VAEGAN-D2~\cite{xian2019fVAEGAND2} & 61.0 & 67.7  \\
        cycle-(U)WGAN~\cite{rafael2018multimodal} & 58.6 & 70.3 \\
        \hline
        Ours & \textbf{66.7} & \textbf{76.8}  \\
        
        \bottomrule
    \end{tabular}
\end{table}

\section{Related Work}
ZSL approaches aim at recognizing objects belonging to classes unseen during training~\cite{larochelle2008zsl,palatucci2009zsl}. This has been extended to the GZSL framework in which the decision space consists of both seen and unseen classes~\cite{socher2013zsl,xian2018zero}. The classical zero-shot approaches build a joint visual-semantic space, relying on a linear cross-modal compatibility function (e.g. dot-product between query embedding and semantic prototypes or a variation of a hinge loss)~\cite{frome2013devise,akata2015evaluation,akata2016label,reed16learning}. Non-linear variants of the compatibility have also been explored~\cite{xian2016latent,socher2013zsl}. Extending previously proposed cross-modal transfer approaches based on auto-encoders~\cite{hubert2017learning} and cross-domain learning~\cite{gretton2007kernel}, more recent line of work~\cite{schonfeld2018generalized,xian2018feature,xian2019fVAEGAND2,rafael2018multimodal,vinay2018generalized} relies on combining these approaches and their variations with dataset augmentation tools such as GAN~\cite{goodfellow2014generative} and VAE~\cite{kingma2014autoencoding}. It is argued that the use of those tools helps to resolve one of the prominent problems in GZSL scenario: classifying images from unseen classes as one of the seen classes. There exist approaches that try to tackle this same problem via temperature calibration~\cite{shichen2018generalized} originally proposed by Hinton et al.~\cite{hinton2015distilling}. Chao et al.\cite{chao2016anempirical} and Das et al.~\cite{das2019zero} proposed approaches to seen/unseen accuracy balancing that are very similar to ours, based on heuristic arguments. We extend this line of work here by providing a probabilistic justification for the balancing effect observed when applying metric rescaling. Atzmon et al.~\cite{atzmon2019adaptive} propose a more sophisticated way to deal with seen/unseen imbalance via adaptive confidence smoothing and gating. In this work, we show that the simpler metric rescaling approach can still be used to achieve impressive results on the GZSL task.  

\begin{figure}[t]
    \centering
    \begin{subfigure}[t]{0.45\linewidth}
        \centering
        \includegraphics[width=\linewidth]{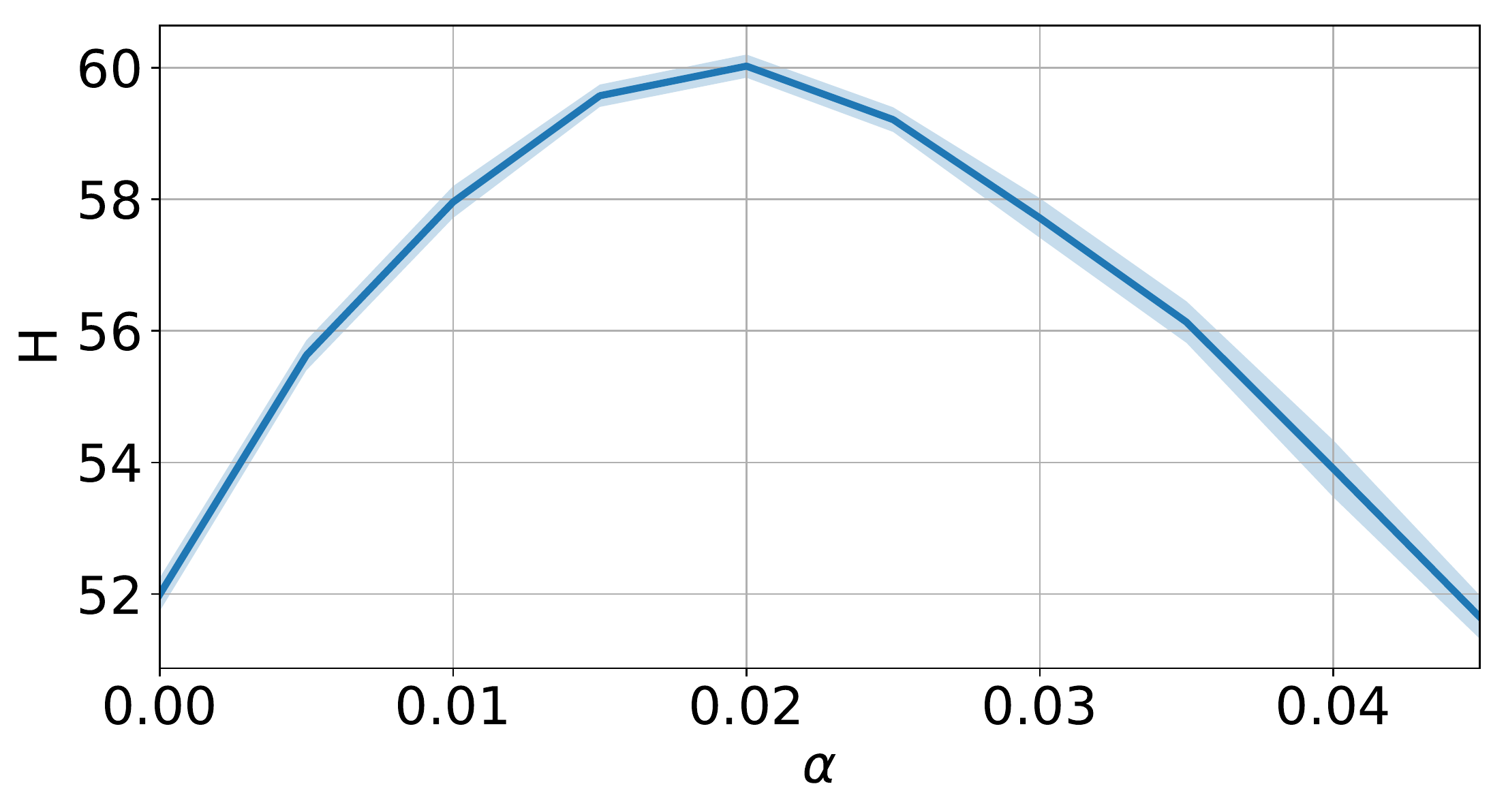}
    \end{subfigure}
    \hspace{0.05\linewidth}
    \begin{subfigure}[t]{0.45\linewidth}
        \centering
        \includegraphics[width=\linewidth]{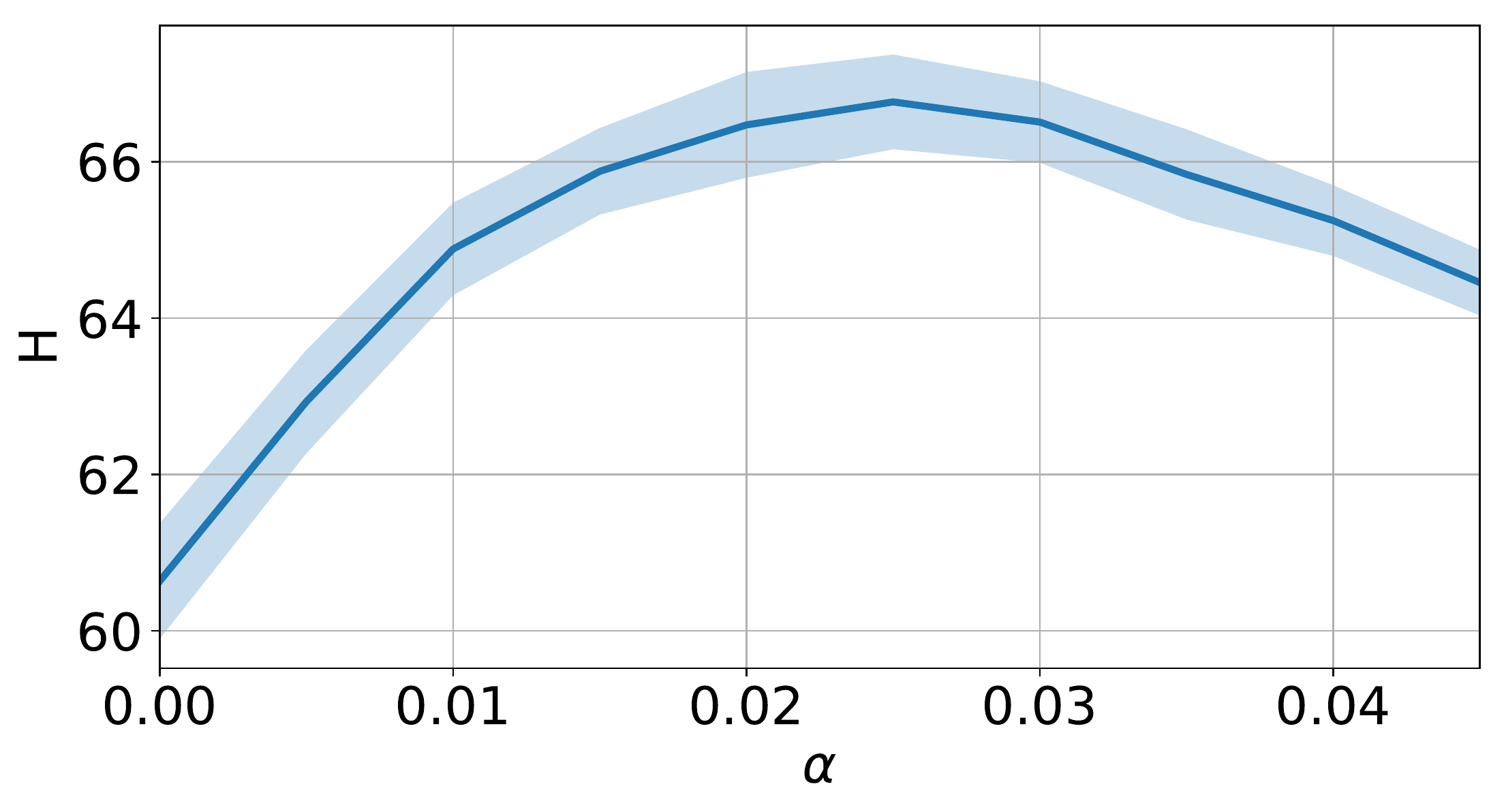}
    \end{subfigure}
    \vspace{-0.1cm}
    \caption{\textbf{H} against $\alpha$ on the validation set, the average and 95\% confidence intervals of 10 repeats. \textbf{H} exhibits a distinct inverted U-shape w.r.t. $\alpha$. \textsc{cub} (left) and \textsc{flowers} (right).}
    \label{fig:crossvalidation_of_alpha}
\end{figure}

\section{Experimental Results}

\textbf{Datasets.} We focus on learning embeddings for fine-grained visual descriptions and test them in ZSL/GZSL scenario. To test the quality of trained embeddings we focus on datasets that provide paired images and text descriptions, such as Caltech-UCSD-Birds (\textsc{cub})~\cite{welinder2010caltech} and Oxford Flowers (\textsc{flowers})~\cite{nilsback2008automated}, that were augmented with textual descriptions by Reed et al.~\cite{reed16learning}. We use the GZSL splits proposed by Xian et al.~\cite{xian2018zero}. The attribute-based datasets, such as SUN~\cite{patterson2014sun} and AWA~\cite{lampert2014attribute} do not contain this information and are out of the scope of the current paper. 

\textbf{Architecture and training details.} see Appendix~\ref{sec:architecture_and_training_details}.
 
\textbf{Our key empirical results} are shown in Tables~\ref{table:key_empirical_result_gzsl} and~\ref{table:key_empirical_result_zsl}. Our results are based on the settings of $\lambda=0.5$, $\kappa=0.5$ and $\alpha$ selected on the validation sets of \textsc{cub} and \textsc{flowers} datasets. Clearly, the combination of the proposed training method and the rebalancing of the metric space results in very impressive performance, especially taking into account the simplicity of our method. In the rest of the section we further analyze the stability with respect to the choices of $\lambda$ and $\kappa$ and provide more details on the selection of $\alpha$. 

\textbf{The seen/unseen accuracy balancing.} Fig.~\ref{fig:crossvalidation_of_alpha}
confirms that \textbf{H} exhibits inverted U-shape behavior as a function of $\alpha$ on the validation sets of \textsc{cub} and \textsc{flowers} datasets, as expected based on results of Section~\ref{ssec:seen_unseen_balancing_theory}. Once the value of $\alpha$ is determined by maximizing \textbf{H} on validation set, we train the representation on the full train+val subset and report results on the test split (the usual practice in GZSL). Validation set construction is detailed in Appendix~\ref{sec:validation_sets}. 

\begin{figure}[t]
    \centering
    \begin{subfigure}[t]{0.45\linewidth}
        \centering
        \includegraphics[width=\linewidth]{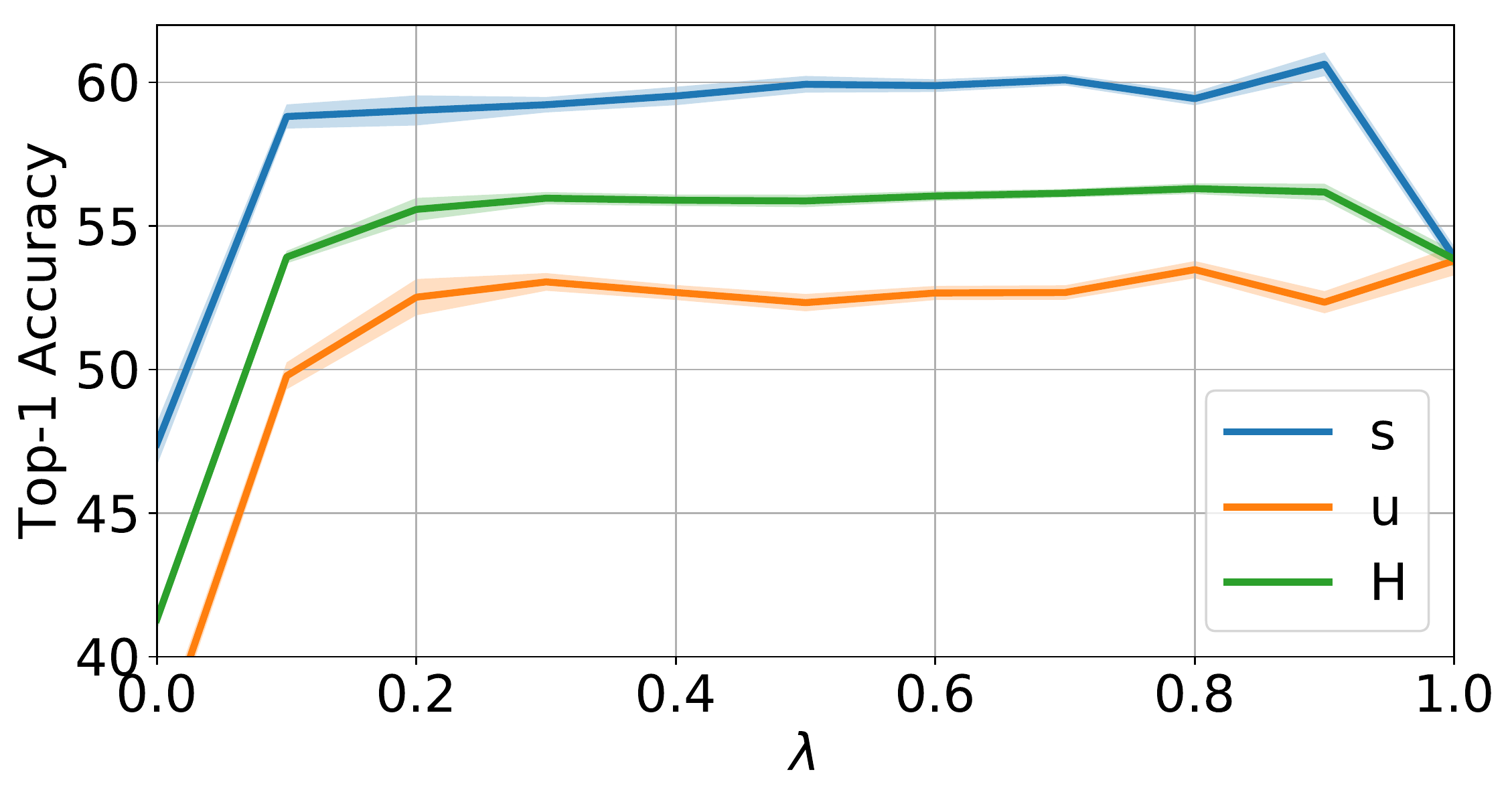}
    \end{subfigure}
    \hspace{0.05\linewidth}
    \begin{subfigure}[t]{0.45\linewidth}
        \centering
        \includegraphics[width=\linewidth]{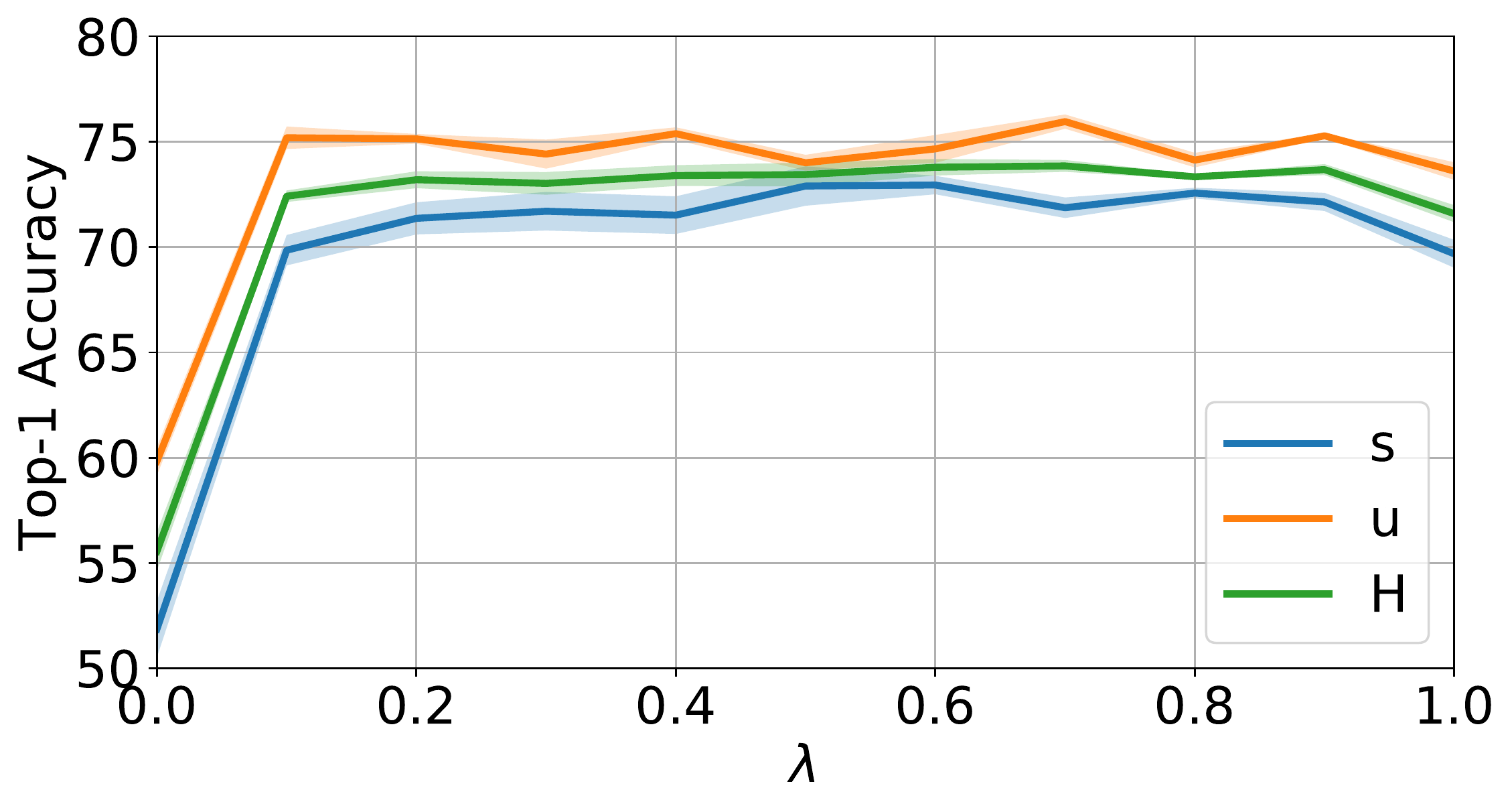}
    \end{subfigure}
    \vspace{-0.1cm}
    \caption{\textbf{H} against $\lambda$, average of 10 repeats. $\lambda=0$ corresponds to the case of disabled text retrieval loss. \textsc{cub} (left) and \textsc{flowers} (right).
    }
    \label{fig:ablation_lambda}
\end{figure}

\begin{table}[t] 
    \centering
    \caption{Generalized zero-shot Top-1 classification accuracy.}
    \label{table:alpha_lambda_kappa_ablation}
    \begin{tabular}{ccc|cccccc} 
        \toprule
        \multicolumn{1}{c}{} & \multicolumn{1}{c}{} & \multicolumn{1}{c}{} & \multicolumn{3}{c}{\textsc{cub}}  & \multicolumn{3}{c}{\textsc{flowers}}  \\ 
        $\alpha$ & $\lambda$ & $\kappa$ & \textbf{u}    &  \textbf{s} & \textbf{H}   & \textbf{u}    &  \textbf{s} & \textbf{H}  \\ \hline
        
        0.0 & 0.5 & 0.5 & 38.3 & 65.3 & 48.3 & 55.1 & 84.6 & 66.7  \\
        
        0.0 & 0.5 & 0.0 & 39.3 & 57.5 & 46.7 & 54.0 & 78.1 & 63.8  \\
        
        \checkmark & 0.5 & 0.0 & 53.8 & 49.6 & 51.6 & 71.7 & 67.2 & 69.4  \\
        
        \checkmark & 0.0 & 0.5 & 47.4 & 36.6 & 41.3 & 51.5 & 60.5 & 55.6  \\
        
        \checkmark & 1.0 & 0.5 & 53.9 & 53.8 & 53.8 & 69.5 & 73.9 & 71.6  \\
        
        \checkmark & 0.5 & 0.5 & 59.3 & 52.6 & \textbf{55.8} & 73.0 & 73.6 & \textbf{73.3}   \\
        \bottomrule 
    \end{tabular}
\end{table}

\begin{figure}[t]
    \centering
    \begin{subfigure}[t]{0.45\linewidth}
        \centering
        \includegraphics[width=\linewidth]{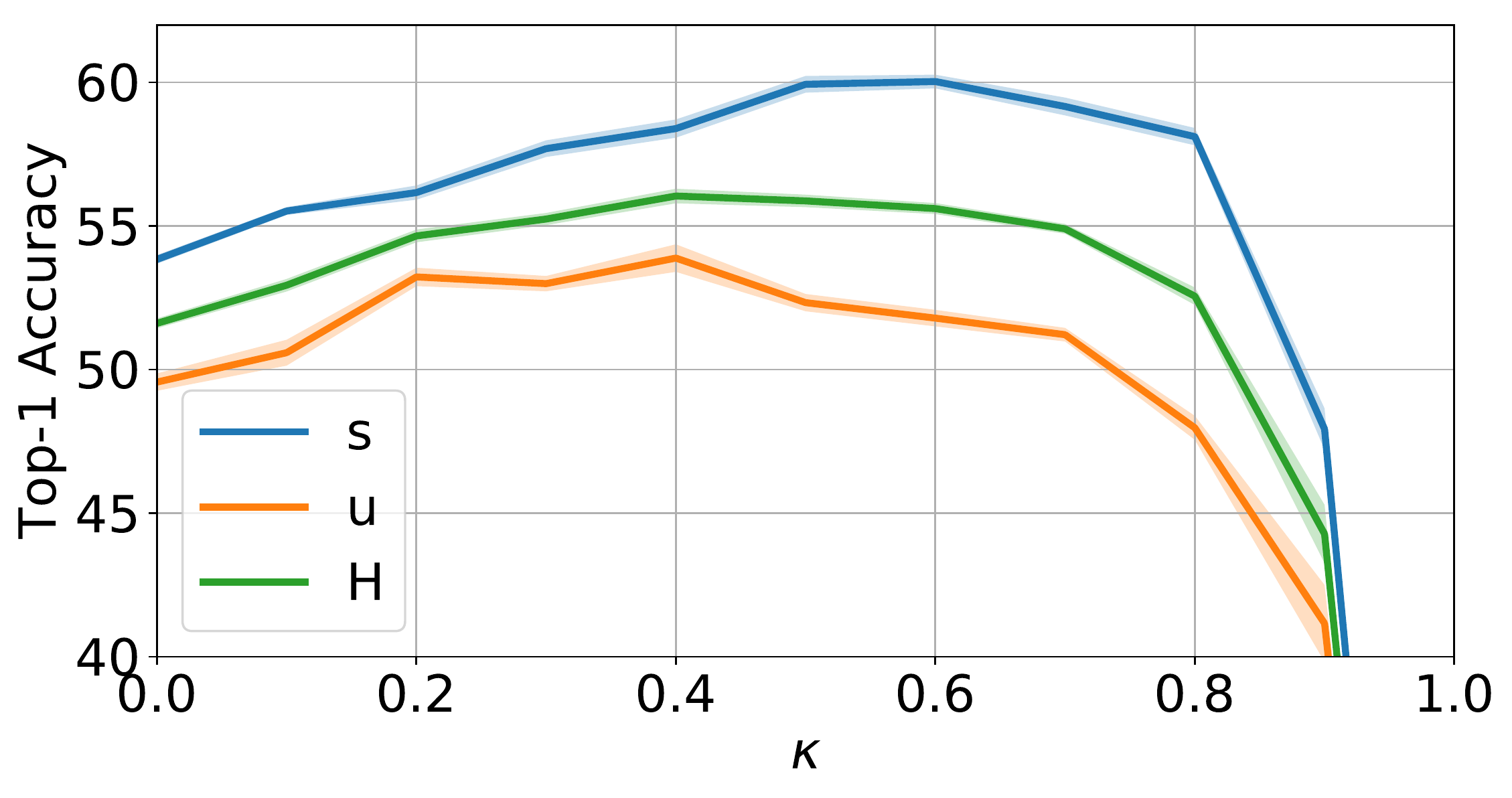}
        \label{fig:ablation_kappa_cub}
    \end{subfigure}
    \hspace{0.05\linewidth}
    \begin{subfigure}[t]{0.45\linewidth}
        \centering
        \includegraphics[width=\linewidth]{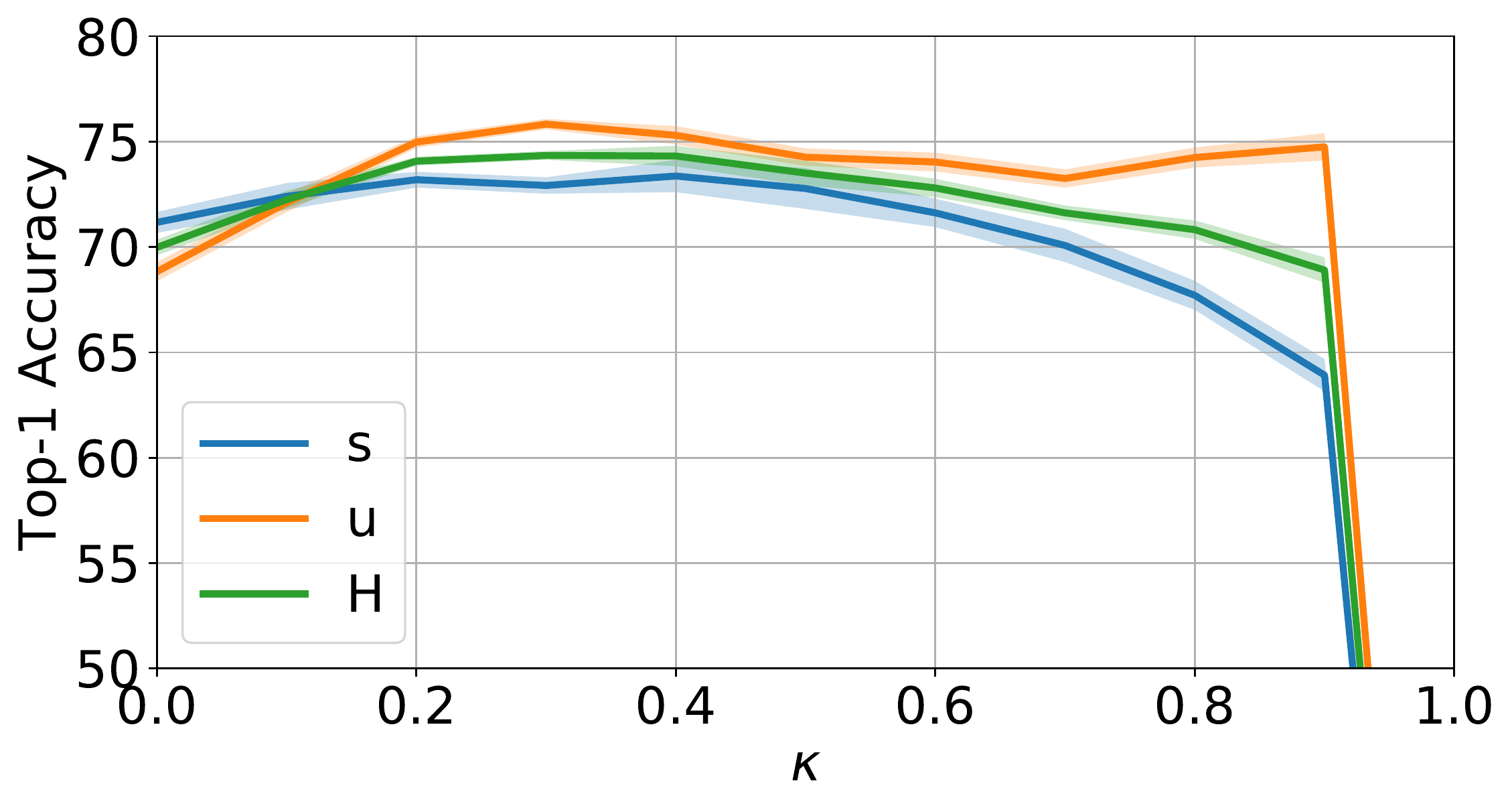}
        \label{fig:ablation_kappa_flowers}
    \end{subfigure}
    \vspace{-0.5cm}
    \caption{\textbf{H} against $\kappa$, average of $10$ repeats. $\kappa=0$ corresponds to the case of classification loss having weight $0$. \textsc{cub} (left) and \textsc{flowers} (right).
    }
    \label{fig:ablation_kappa}
\end{figure}

\textbf{Ablation studies.} Fig.~\ref{fig:ablation_lambda} studies the importance of image and text retrieval losses. We see that all Top-1 accuracies (\textbf{H}, \textbf{s}, \textbf{u}) are stable in the range $\lambda \in [0.2, 0.9]$. Removing text retrieval loss ($\lambda=0$) results in the most significant drop. Indeed, at the batch level, retrieving the correct text given an image is related to identifying the correct class encoded by a text prototype during GZSL inference step. Fig.~\ref{fig:ablation_kappa} studies the interplay between the retrieval and the classification losses. We again observe that there exists a reasonably stable range of $\kappa \in [0.2, 0.6]$. $\kappa=1$ results in the catastrophic performance drop: the classification losses alone do not enforce the necessary modality alignment.

Table~\ref{table:alpha_lambda_kappa_ablation} further studies the effects of different loss terms. The best result is achieved when all loss terms are active and when the metric rescaling is on (the last line in the table: $\lambda,\kappa=0.5$ and $\alpha$ checked). Comparing this to the case with no metric rescaling (first line, $\alpha=0$), we see that the rescaling helps to greatly decrease the gap between seen and unseen classification accuracy, both on \textsc{cub} and \textsc{flowers}. Interestingly, we only use images and texts from the training set to achieve it. Going to the second line in the table (the image/text classification loss is inactive, $\kappa=0$) and comparing it to the first one, we assess the effect of the image/text classification loss. It barely affects the performance on unseen set, but it significantly boosts the classification accuracy on the seen set (around 8\% on both datasets). However, it improves GZSL accuracy only when applied together with metric rescaling (please refer to lines 1 and 6 in Table~\ref{table:alpha_lambda_kappa_ablation}). Our interpretation is that the image/text classifier loss reduces the intraclass variability and enforces tighter  embedding clustering. Yet, this also leads to overfit on classification task. This is accounted for by metric rescaling that enables the learnings from the image/text classification task be transferred effectively into the GZSL task. Finally, an interesting observation can be made by comparing line 3 of Table~\ref{table:alpha_lambda_kappa_ablation} with performance of algorithms in Table~\ref{table:key_empirical_result_gzsl}. In this case our training relies only on retrieval losses computed without class labels solely based on the pairwise relationships between texts and images. The learned representation is competitive against the latest GAN/VAE based approaches on \textsc{cub} and is state-of-the-art on \textsc{flowers}. We conclude that when very fine-grained modality outputs are available (image and text pairs being a very prominent example), the high-quality representations may be learned without relying on manually supplied class labels.

\section{Conclusions}

We propose and empirically validate two contributions for learning fine-grained cross-modal representations. First, we confirm the hypothesis that in the context of paired images and texts, a deep metric learning approach can be driven by an instance-based retrieval loss resulting in impressive GZSL classification results. This demonstrates that high-quality deep representations can be trained relying largely on pairwise modality relationships. Second, we mathematically analyze and empirically validate a simple method of balancing seen/unseen accuracy in the GZSL task.

\clearpage
{\small
\bibliographystyle{ieee_fullname}
\bibliography{main}
}

\clearpage
\appendix

\section{Loss calculation algorithm} \label{sec:loss_algorithm}

\begin{algorithm*}[t] 
    \caption{Loss calculation for a single optimization iteration of the proposed method. $N$ is the number of instances in the training set $\trainset{tr}$, $B$ is the number of instances per batch, $C$ is the number of classes in the train set. $\textsc{RandomSample}(\mathcal{S}, B)$ denotes a set of $B$ elements chosen uniformly at random from a set $\mathcal{S}$, without replacement.}
    \label{alg:one_iteration}
    \begin{algorithmic}
    \REQUIRE Training set $\trainset{tr} = \{ (v_1, t_1, y_1), \ldots, (v_N, t_N, y_N) \}$, $\lambda \in [0,1]$, $\kappa \in [0,1]$. 
    \ENSURE The loss $J(\phi, \theta)$ for a randomly sampled training batch.
    \STATE $\mathcal{I} \gets \textsc{RandomSample}(\{1, \ldots, N\}, B)$ 
        \COMMENT{Select $B$ instance indices for batch}
    \STATE $J_{TC}(\theta), J_{IC}(\phi) \gets 0, 0$ \COMMENT{Initialize classification losses}
    \FOR{$i$ in $\mathcal{I}$}
        \STATE $\vec{z}_{v_i}, \vec{z}_{t_i} \gets f_{\phi}(v_i)$, $f_{\theta}(t_i)$ \COMMENT{Embed images and texts}
        \STATE $p_{I} \gets \softmax(\vec{W}_{I} \vec{z}_{v_i} + \vec{b}_{I})$ \COMMENT{Image classifier probabilities}
        \STATE $p_{T} \gets \softmax(\vec{W}_{T} \vec{z}_{t_i} + \vec{b}_{T})$
        \COMMENT{Text classifier probabilities}
        \STATE $J_{TC}(\theta) \gets J_{TC}(\theta) + \frac{1}{B} \crossentropy(p_{T}, y_i)$ \COMMENT{Text classification loss}
         \STATE $J_{IC}(\phi) \gets J_{IC}(\phi) + \frac{1}{B} \crossentropy(p_{I}, y_i)$ \COMMENT{Image classification loss}
    \ENDFOR
    \STATE $J_{TR}(\phi, \theta), J_{IR}(\phi, \theta) \gets 0, 0$ \COMMENT{Initialize retrieval losses}
    \FOR{$i$ in $\mathcal{I}$}
        \STATE $\displaystyle J_{TR}(\phi, \theta) \gets J_{TR}(\phi, \theta) + 
        \frac{1}{B} \left[
            d(\vec{z}_{v_i}, \vec{z}_{t_i}) + \log \sum_{j \in \mathcal{I}} \exp(-d(\vec{z}_{v_i}, \vec{z}_{t_j}))
        \right]$ \COMMENT{Text retrieval loss}
        \STATE $\displaystyle J_{IR}(\phi, \theta) \gets J_{IR}(\phi, \theta) + 
        \frac{1}{B} \left[
            d(\vec{z}_{v_i}, \vec{z}_{t_i}) + \log \sum_{j \in \mathcal{I}} \exp(-d(\vec{z}_{t_i}, \vec{z}_{v_j}))
        \right]$ \COMMENT{Image retrieval loss}
    \ENDFOR
    
    \STATE $J(\phi, \theta) \gets \lambda J_{TR}(\phi, \theta) + (1-\lambda)J_{IR}(\phi, \theta)$ \COMMENT{Add retrieval loss to the total loss}
    \STATE $J(\phi, \theta) \gets (1-\kappa) J(\phi, \theta) + \frac{\kappa}{2} (J_{TC}(\theta) + J_{IC}(\phi))$ \COMMENT{Add classification loss to the total loss}
    \end{algorithmic}
\end{algorithm*}

\section{The Analysis of Error Rates} \label{sec:proof}
First define the misclassification of unseen as seen classes for the classifier $\widehat y_{\alpha}$, based on $d_{\alpha}(\vec{z}_v, \vec{p}(y))$:
\begin{align} \label{eqn:seen_unseen_probability_representation_scaled}
\Proba\{ \widehat y_{\alpha} &\in \Y{tr}{} | y_v \in \Y{ts}{} \} = \Proba\left\{(1+\alpha)\min_{y \in \Y{tr}{}} d(\vec{z}_v, \vec{p}(y)) \right. \nonumber \\
&< \left. \min_{y \in \Y{ts}{}} d(\vec{z}_v, \vec{p}(y)) | y_v \in \Y{ts}{} \right\}\;,
\end{align}
We show that $\Proba\{ \widehat y \in \Y{tr}{} | y_v \in \Y{ts}{} \} \geq \Proba\{ \widehat y_{\alpha} \in \Y{tr}{} | y_v \in \Y{ts}{} \}$. 
Let us define $\delta_{tr} \equiv \min_{y \in \Y{tr}{}} d(\vec{z}_v, \vec{p}(y))$ and $\delta_{ts} \equiv \min_{y \in \Y{ts}{}} d(\vec{z}_v, \vec{p}(y))$, then Equation~\eqref{eqn:seen_unseen_probability_representation_scaled} can be rewritten as:
\begin{align} 
\Proba\{ \widehat y_{\alpha} \in \Y{tr}{} | y_v \in \Y{ts}{} \} = \Proba\left\{(1+\alpha)\delta_{tr} < \delta_{ts} | y_v \in \Y{ts}{} \right\}\;.
\end{align}
Let us consider the probability of event $\delta_{tr} < \delta_{ts}$ and decompose it as follows:
\begin{align*} 
&\Proba\left\{\delta_{tr} < \delta_{ts} | y_v \in \Y{ts}{} \right\} = \Proba\left\{(1+\alpha)\delta_{tr} < (1+\alpha)\delta_{ts} | y_v \in \Y{ts}{}  \right\} \\
&= \Proba\left\{(1+\alpha)\delta_{tr} < \delta_{ts} \cup \delta_{ts} \leq (1+\alpha)\delta_{tr} < (1+\alpha)\delta_{ts} | y_v \in \Y{ts}{}  \right\} \\
&= \Proba\left\{(1+\alpha)\delta_{tr} < \delta_{ts} | y_v \in \Y{ts}{} \right\} + \Proba\left\{ \delta_{ts} \leq (1+\alpha)\delta_{tr} < (1+\alpha)\delta_{ts} | y_v \in \Y{ts}{}  \right\} \\
&- \Proba\left\{(1+\alpha)\delta_{tr} < \delta_{ts} \cap \delta_{ts} \leq (1+\alpha)\delta_{tr} < (1+\alpha)\delta_{ts} | y_v \in \Y{ts}{}  \right\} \\
&= \Proba\left\{(1+\alpha)\delta_{tr} < \delta_{ts} | y_v \in \Y{ts}{}\right\} + \Proba\left\{ \delta_{ts} \leq (1+\alpha)\delta_{tr} < (1+\alpha)\delta_{ts} | y_v \in \Y{ts}{}  \right\}\;.
\end{align*}
The transitions are based on the relationship between probabilities of arbitrary events $A$ and $B$, $\Proba\{A \cup B\} = \Proba\{ A \} + \Proba\{ B\} - \Proba\{ A \cap B \}$, and in our case $\Proba\{ A \cap B \} = 0$. This implies that:
\begin{align} 
\Proba\{ \widehat y_{\alpha} \in \Y{tr}{} | y_v \in \Y{ts}{} \} &= \Proba\{ \widehat y \in \Y{tr}{} | y_v \in \Y{ts}{} \} - \Proba\left\{ \frac{\delta_{ts}}{(1+\alpha)} \leq \delta_{tr} < \delta_{ts} | y_v \in \Y{ts}{}  \right\} \nonumber \\ 
&\leq \Proba\{ \widehat y \in \Y{tr}{} | y_v \in \Y{ts}{} \}.
\end{align}

We have just shown that for a non-negative $\alpha$ the probability of misclassifying an image from an unseen class as one of the seen classes is smaller for the decision rule $\widehat y_{\alpha}$ than for the original decision rule $\widehat y$. In fact, we can make a stronger claim. Since $\delta_{ts}$ and $\delta_{tr}$ are non-negative, it is clear that the length of interval $\left[ \delta_{ts}/(1+\alpha),  \delta_{ts} \right)$ increases as $\alpha$ increases, and hence probability that $\delta_{tr}$ falls in this interval is non-decreasing with increasing $\alpha$. Thus we have for any $0 \leq \alpha_1 \leq \alpha_2$, $\Proba\{ \widehat y_{\alpha_1} \in \Y{tr}{} | y_v \in \Y{ts}{} \} \geq \Proba\{ \widehat y_{\alpha_2} \in \Y{tr}{} | y_v \in \Y{ts}{} \}$, \emph{i.e.} $\Proba\{ \widehat y_{\alpha} \in \Y{tr}{} | y_v \in \Y{ts}{} \}$ is a monotone non-increasing function of $\alpha$ and we can reduce it by increasing $\alpha$.

\clearpage
\section{Constructing validation sets} \label{sec:validation_sets}

The validation set is constructed by further splitting the train set on \textsc{cub} and \textsc{flowers}. For example, \textsc{cub} has a train set of 5875 images from 100 seen classes and a validation set of 2946 images from 50 unseen classes. We further divide the train set into 4700 train images from 100 seen classes, 1175 seen validation images (4700 + 1175 = 5875) and we use all the 2946 images from 50 classes as the unseen validation set.

\section{Architecture and Training Details} \label{sec:architecture_and_training_details}

The text feature extractor is built by cascading two residual CNN blocks, followed by a BiLSTM. Each block has 3 convolutional/batch norm layers. The number of filters in the blocks is 128 and 256, BiLSTM has 512 filters for forward and backward branches (1024 total). All variables in the convolutional stack (including the batch normalization parameters $\gamma$ and $\beta$) are L2-penalized with weight $0.001$. The image feature extractor is a ResNet-101 with  fixed weights pretrained on the split of ImageNet proposed by Xian et al.~\cite{xian2018zero}. In this work we use precomputed image features, available in~\cite{xian2017cubfeatures} for \textsc{cub} and in~\cite{xianCVPR18supplementary} for \textsc{flowers}. Image and text features are projected in the common embedding space of size 1024 with FC layers and no non-linearity. They are preceded with a dropout of 0.25. The trainable components of the model are trained for 150k batches of size 32 using SGD with initial learning rate of $0.1$ that is annealed by a factor of 10 every 50k batches. For each batch, we sample 32 instances, each instance includes a vector of precomputed ResNet-101 features and 10 text descriptions corresponding to it, according to the original dataset definition~\cite{reed16learning}. All 10 text descriptions are processed via the CNN/LSTM stack and the resulting embeddings are average pooled to create a vector representation of length 1024.  

\end{document}